\documentclass{article}

\usepackage{float}
\usepackage{booktabs}
\usepackage{amsmath}
\usepackage{subfigure}
\usepackage{graphicx}
\usepackage{makecell}
\usepackage{multicol}
\usepackage{multirow}
\usepackage{color,soul}
\usepackage{wrapfig}       
\usepackage{adjustbox}     

\usepackage{amsmath,amsfonts,bm}

\def\eqref#1{equation~\ref{#1}}

\def\1{\bm{1}}

\DeclareMathAlphabet{\mathsfit}{\encodingdefault}{\sfdefault}{m}{sl}
\SetMathAlphabet{\mathsfit}{bold}{\encodingdefault}{\sfdefault}{bx}{n}

\DeclareMathOperator*{\argmin}{arg\,min}

\usepackage[final]{neurips_fitml_2024}

\usepackage[utf8]{inputenc} 
\usepackage[T1]{fontenc}    
\usepackage{hyperref}       
\usepackage{url}            
\usepackage{booktabs}       
\usepackage{amsfonts}       
\usepackage{nicefrac}       
\usepackage{microtype}      
\usepackage{xcolor}         
\usepackage{bbm}

\title{Ensembling Finetuned Language Models\\ for Text Classification}

\author{Sebastian Pineda Arango$^1$\thanks{Equal contribution. Corresponding author: \texttt{pineda@cs.uni-freiburg.de}} , Maciej Janowski$^1$\footnotemark[1] , Lennart Purucker$^1$, Arber Zela$^1$, \\
    \textbf{Frank Hutter$^{3,1}$,\ Josif Grabocka$^{2}$}
    \vspace*{2mm}\\
    $^1$ University of Freiburg, $^2$ University of Technology Nürnberg, $^3$ ELLIS Institute Tübingen
}

\begin{document}

\maketitle

\begin{abstract}
Finetuning is a common practice widespread across different communities to adapt pretrained models to particular tasks.
Text classification is one of these tasks for which many pretrained models are available. On the other hand, ensembles of neural networks are typically used to boost performance and provide reliable uncertainty estimates. However, ensembling pretrained models for text classification is not a well-studied avenue. 
In this paper, we present a metadataset with predictions from five large finetuned models on six datasets, and report results of different ensembling strategies from these predictions. Our results shed light on how ensembling can improve the performance of finetuned text classifiers and incentivize future adoption of ensembles in such tasks.

\end{abstract}

\section{Introduction}

In recent years, fine-tuning pretrained models has become a widely adopted technique for adapting general-purpose models to specific tasks~\citep{arango2023quick}. This practice has gained significant traction across various communities due to its effectiveness in leveraging the vast knowledge encoded in pretrained models. Among the diverse tasks that benefit from fine-tuning, text classification stands out as one of the most prevalent. With the availability of numerous pretrained models, practitioners often find themselves with a range of powerful tools to tackle text classification challenges.
However, despite the widespread use of fine-tuning, the potential benefits of combining or ensembling these fine-tuned models remain underexplored.

Previous studies have primarily concentrated on improving individual model performance through fine-tuning techniques~\citep{ruder2018universal}, leaving the exploration of ensemble strategies largely underdeveloped in this context. This oversight is particularly significant given the well-documented advantages of model ensembling in other machine learning domains~\citep{agtabular,lakshminarayanan2017simple}, which has been shown to enhance robustness and generalization. In this paper, we address the aforementioned gap by introducing a novel metadataset, which we dub: Finetuning Text Classifiers (FTC) metadataset. FTC contains predictions from various fine-tuned models on text classification tasks with various number of classes. We systematically evaluate different ensembling strategies using this metadataset, aiming to uncover insights into the potential improvements that ensembling can offer. Our results provide valuable evidence on the efficacy of these strategies, demonstrating that ensembling fine-tuned models can lead to performance gains in text classification. 



\section{Background and Related Work}

\textbf{Finetuning for Text Classification.} Universal Language Model Fine-tuning for Text Classification or ULMFiT~\citep{ruder2018universal} consists of finetuning language models for classification in two stages: 1) a target task unsupervised finetuning and 2) target task classifier finetuning, while using a different learning rate per layer. However, the feasibility of fully fine-tuning large pretrained language models is constrained by computational limits~\citep{radford2018plm}. This has spurred the adoption of Parameter-Efficient Fine-Tuning (PEFT) methods~\citep{han2024peftsurvey}. Early strategies focused on minimal subsets of parameters such as sparse subnetworks~\citep{ sung2021subnets} to improve task-specific performance efficiently. Innovations such as adapter modules~\citep{houlsby19adapters}, which introduce a few parameters per transformer layer but in consequence increase inference time, prompted the development of Low-Rank Adaptation (LoRA)~\citep{hu2022lora, dettmers2024qlora} that applies low-rank updates for improved downstream task performance with reduced computational overhead. Some studies have also demonstrated that finetuned language models can be ensembled to improve performance for text classification~\citep{abduri2027generative}, but they do not provide clear insights about ensembling methods, hyperparameters, or metadata.




\textbf{Ensembling Deep Learning Models.}
Ensembles of neural networks~\citep{hansen_ensembles, krogh-ens, dietterich_ensemble_methods} have gained significant attention in deep learning research, both for their performance-boosting capabilities and their effectiveness in uncertainty estimation. Various strategies for building ensembles exist, with deep ensembles~\citep{lakshminarayanan2017simple} being the most popular one, which involve independently training multiple initializations of the same network.
Their state-of-the-art predictive uncertainty estimates have further fueled the interest in ensembles. Extensive empirical studies~\citep{ovadia19, gustafsson2019evaluating} have shown that deep ensembles outperform other approaches for uncertainty estimation, such as Bayesian neural networks~\citep{pmlr-v37-blundell15, pmlr-v48-gal16, sgld11}. Similar to our work, \citet{beyondDE_23} show that finetuning pretrained models via Bayesian methods on the WILDS dataset~\citep{wilds-koh21a}, which contains text classification as well, can yield significant performance as compared to standard finetuning of single models.

\textbf{Post-Hoc Ensembling (PHE).}
PHE uses set of fitted base models $\{z_1,...,z_M\}$ such that every model outputs $z_m(x), z_m: \mathbb{R}^D \rightarrow \mathbb{R}^C$~\footnote{We assume a classification tasks with $C$ classes. For regression $C=1$.}. These outputs are combined by an ensembler $f(z_1(x), ..., z_M(x) ; \theta)=f(z(x); \theta)$, where $z(x) = [z_1(x),...,z_M(x)]$ is the concatenation of the base models predictions. While the base models learned from a training set $\mathcal{D}_{\mathrm{Train}}$, the ensembler's parameters $\theta$ are typically obtained by minimizing a loss function $\mathcal{L}$ on a validation set $\mathcal{D}_{\mathrm{Val}}$ such that:
\begin{equation}
    \theta \in \argmin_{\theta} \sum_{(x,y) \in \mathcal{D}_{\mathrm{Val}}}\mathcal{L}(f(z(x), y; \theta)).
    \label{eq:post-hoc-ensembling}
\end{equation}
A popular approach is a linear combination of the model outputs as $f(z(x); \theta) =  \sum_m \theta_m z_m(x)$. 


\textbf{PHE Metadatasets.} Similarly, prior studies have created metadatasets containing the \textit{predictions} of base models, but only for time-series \citep{borchert2022multi} and tabular \citep{purucker2022assembled,purucker2023cma, purucker2023q,salinas2023tabrepo} data.

\section{Finetuning Text Classifiers (FTC) Metadataset }

\begin{wraptable}{R}{.69\textwidth}
    \vspace{-1.5em}
    \centering
        \caption{Search Space parameterization.}
        \resizebox{\linewidth}{!}{%
    \begin{tabular}{lc}
    \toprule
    \bfseries Hyperparameter & \bfseries Values  \\
     \midrule
     Model & GPT2, Bert-Large, Albert-Large, Bart-Large, T5-Large \\
     Learning Rate & $0.00001$, $0.0001$, $0.0005$, $0.001$, $0.005$\\
     LoRA Rank & $8$, $16$, $32$, $64$, $128$\\
     \bottomrule
\end{tabular}
    }
    \label{tab:search_space}
\end{wraptable}

\textbf{Search Space.} Our search space comprises three hyperparameters: the model type, learning rate and LoRA rank~\citep{hu2022lora}.  We consider five model choices: 1) \textbf{GPT2, 124M} parameters;~\citep{radford2019language}; 2) \textbf{Bert-Large, 336M}  ;~\citep{DBLP:journals/corr/abs-1810-04805}; 3) \textbf{Bart-Large, 400 M}, parameters ~\citep{bart}; 4) \textbf{Albert-Large, 17M} parameters ~\citep{albert}; and 5) \textbf{T5-Large, 770 M} parameters ~\citep
{2020t5}. For the other two hyperparameters we also consider five different discrete values as specified in Table \ref{tab:search_space}.


\begin{table}[t]
      \caption{Metadataset information.}
    \centering
    \resizebox{\textwidth}{!}{%
    \begin{tabular}{lccccccc}
    \toprule
    \bfseries Dataset & \bfseries \# Classes & \bfseries \# Train Samples & \bfseries \#  Val. Samples & \bfseries \#  Test Samples & \bfseries \# Confs (100\%) & \bfseries \# Confs. (10\%) \\
     \midrule
     IMDB~\citep{imdb_dataset}  & $2$ & $20,000$ & $5,000$ & $25,000$ & $125$ & $125$\\
     Tweet~\citep{tweet-sentiment-extraction} & $3$ & $27,485$ & $5,497$ & $3,534$ & $100$ & $100$\\
     News~\citep{Zhang2015CharacterlevelCN} & $4$ & $96,000$ & $24,000$ & $7,600$ & $99$ & $120$\\
     DBpedia~\citep{Zhang2015CharacterlevelCN} & $14$ & $448,000$ & $112,000$ & $70,000$ & $25$ & $65$\\
     SST-2  ~\citep{socher-etal-2013-recursive}&  $2$ & $43,103$ & $13,470$ & $10,776$ & $125$ & $125$\\
     SetFit ~\citep{setfit-mnli} & $3$ & $393,116$ & $78,541$ & $62,833$ & $25$ & $100$\\

     \bottomrule
\end{tabular}
    }
    \label{tab:metadataset_information}
\end{table}


\textbf{Datasets.} The metadataset contains predictions of models finetuned on five metadatasets for text classification: 1) \textit{IMDB}~\citep{imdb_dataset}; 2) \textit{Tweet}~\citep{tweet-sentiment-extraction}, 3) \textit{News}~\citep{Zhang2015CharacterlevelCN}, 4) \textit{DBpedia}~\citep{Zhang2015CharacterlevelCN}, 5) \textit{SST2}~\citep{socher-etal-2013-recursive} and 6) \textit{SetFit}~\citep{setfit-mnli}.  We created two versions for every dataset: the first is trained with the complete training data, while the second is only with a subset of $10\%$ of the samples. All the datasets are for text classification from $2$ to $14$ classes, including diverse domains such as movies, reviews, news, tweets, and text entailment data. We provide further information on the datasets in Table~\ref{tab:metadataset_information} and Appendix \ref{apx:dataset_information}.

\begin{wraptable}[8]{R}{.65\textwidth}
    \vspace{-1.7em}
    \centering
        \caption{Best configuration per dataset.}
        \resizebox{\linewidth}{!}{%
    
\begin{tabular}{lcccccc}
\toprule
\multirow{2}[3]{*}{} & \multicolumn{3}{c}{\bfseries 100 \%} & \multicolumn{3}{c}{\bfseries 10 \%} \\ \cmidrule(lr){2-4} \cmidrule(lr){5-7} 
     \bfseries Dataset &  \bfseries Model &  \bfseries Learning Rate &  \bfseries Lora Rank &  \bfseries Model &  \bfseries Learning Rate &  \bfseries Lora Rank  \\\midrule
     \bfseries DBpedia & GPT2 & $0.0001$ & $64$ & Bert & $0.0001$ & $16$ \\
     \bfseries News & Bart & $0.0001$ & $64$ & Bart & $0.0001$ & $128$ \\
     \bfseries SetFit & GPT2 & $0.0001$ & $128$ & Bart  & $0.0001$ & $8$\\
     \bfseries SST2 & T5 & $0.0001$ & $8$ & T5 & $0.0001$ & $64$\\
     \bfseries Tweet & Bart & $0.0001$ & $64$ & Bart & $0.0001$ & $64$\\
     \bfseries IMDB & Bart & $0.0001$ & $128$ & T5 &  $0.0001$ & $64$\\
\bottomrule
\end{tabular}

    }
    \label{tab:my_label}
\end{wraptable}

\textbf{Metadataset Creation and Composition.\footnote{Access to the metadataset and finetuning code in \url{https://github.com/sebastianpinedaar/finetuning_text_classifiers}}} We created the dataset by finetuning every model to the train split and, subsequently, saving their predictions on the validation and test split. This allows us to quickly simulate ensembling methods given the precomputed predictions. The validation split corresponds to $20\%$ of the available train data. For \textit{SST-2} and \textit{SetFit} the test data is not completely provided by the creators, or it has hidden labels, therefore, we obtain it by using $20\%$ of the remaining training data. The models are finetuned up to $5$ epochs using a single Nvidia A100 GPU with batch size set to $2$ and no LoRA dropout. We vary only the model type, learning rate, and LoRA rank, while keeping the other hyperparameters to their default values in the \textsc{Trainer} object from the \textit{Transformers Library} (v4.41.0)~\footnote{Although we evaluate the models in a grid, some runs yielded out-of-memory errors for some configurations.}.
In total, the metadataset contains $1134$ evaluated configurations, representing around $3800$ GPU hours of computation. Additionally, we report information about the metadataset in Table \ref{tab:metadataset_information}, as well as training times per dataset in Table \ref{tab:times}.

\begin{wrapfigure}[16]{R}{.35\textwidth}
\vspace{-1em}
\includegraphics[width=0.9 \linewidth]{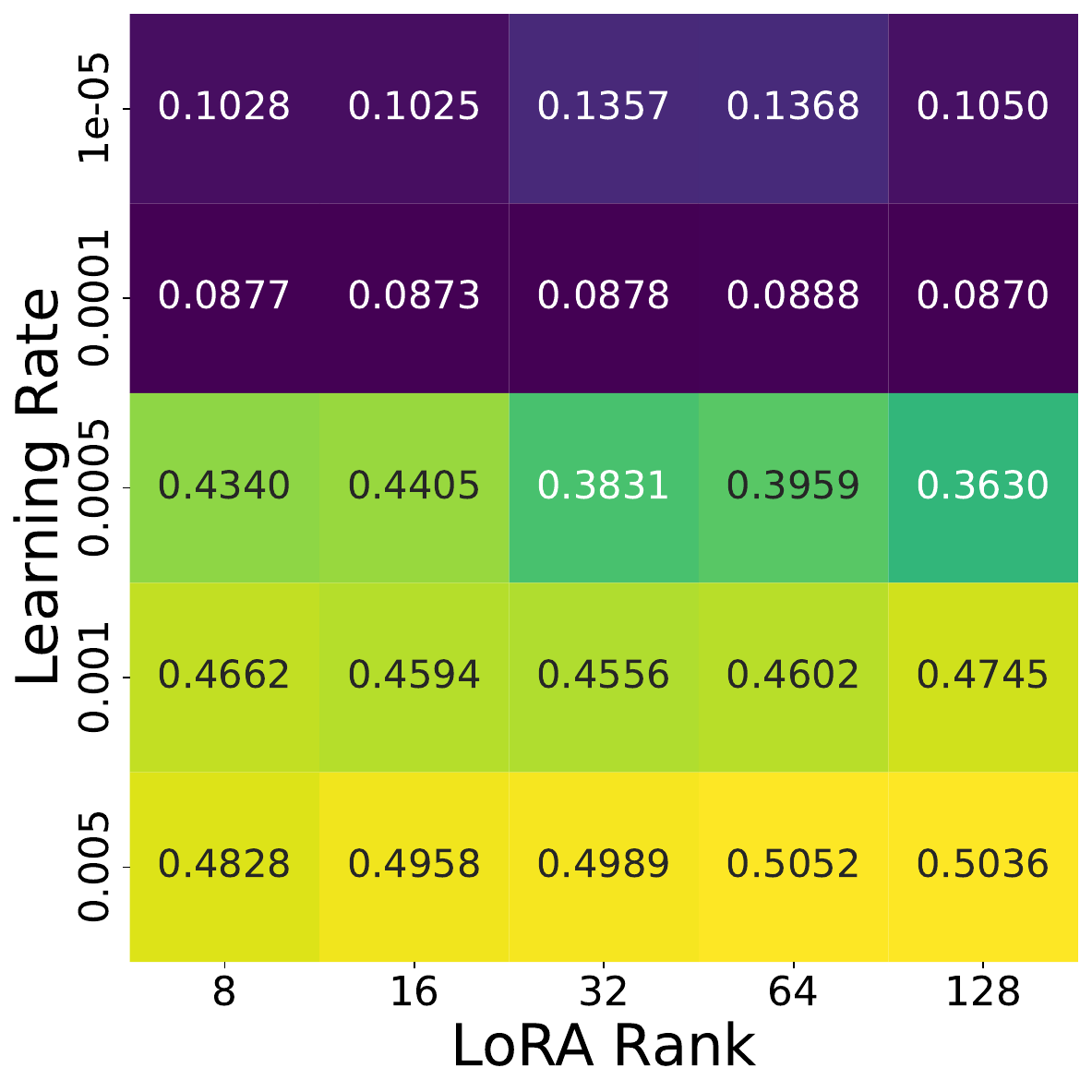} 
\caption{Mean error across datasets for different hyperparameter combinations. }
\label{fig:error_hyper_extended}
\end{wrapfigure}

\textbf{Hyperparameter Importance.}
We explore the importance of two hyperparameters, learning rate, and LoRA rank, by plotting the mean error as a heatmap in Figure~\ref{fig:error_hyper_extended}. The error corresponds to the average across different models and datasets. We can notice that the learning rate is an important hyperparameter, while increasing the LoRA rank does not affect the performance significantly in the low learning rate regime. This behaviour is interesting, as it showcases that a small rank is enough for successful finetuning in this context. A similar pattern arises when using $10\%$ of the data, as shown in Figure \ref{fig:error_hyper_mini} in the appendix. To compare the different classifiers, we report their test error on all dataset versions after selecting the best LoRA rank and learning rate configuration, in Table \ref{tab:error_per_mode}. T5-large shows strong performance in comparison to the other models. Bart and GPT2 also outperform the rest of the models in some datasets. These results demonstrate that the model type is also a relevant hyperparameter, which might motivate the exploration of joint model/architecture and hyperparameter optimization for achieving the best performance.


\section{Benchmarking Ensembles of Finetuned Text Classifiers}

We compare the Neural Ensemblers with other common and competitive ensemble approaches. 1) \textbf{Single best} selects the best model according to the validation metric; 2) \textbf{Random-N} chooses randomly $N$ models to ensemble, 3) \textbf{Top-N} ensembles the best $N$ models according to the validation metric; 4) \textbf{Greedy-N} creates an ensemble with $N$ models by iterative selecting the one that improves the metric as proposed by previous work~\citep{caruana2004ensemble,caruana2006getting}; 5) \textbf{Model Average (MA)} simply computes the sum of the predictions with constant weights.
For some baselines, we tried both $5$ and $50$ models in the ensembles, e.g. \textit{Greedy-50} has $50$ models.

\subsection{Observation 1: Ensembling finetuned text classifiers is helpful.}

To understand whether it is helpful to ensemble finetunined text classifier, we evaluate the baselines on the six datasets, on both versions with $100\%$ and $10\%$ of the training data. We measure the negative log-likelihood (NLL) and the classification error on the test data, while we use the validation split for training the ensemble.
From results shown in Tables \ref{tab:error} and \ref{tab:nll}, we observe that the best method (bold-faced) is always an ensembling technique. Except Random-N, all the other ensembling strategies yield consistently better results than the single-best approach, which corresponds to a grid search on the search space of configurations. Particularly, we notice that the Greedy-N approach is very strong across all datasets, especially regarding the NLL. A large ensemble ($50$ base models) seems to be beneficial using the \textit{Greedy-N} approach, but the results are mixed when using the \textit{Top-50} or \textit{Random-50}. We notice a particular large improvement in the NLL metric, confirming that ensembling provides robustness and better uncertainty calibrations~\citep{lakshminarayanan2017simple, ovadia19, beyondDE_23}.

\begin{table}[]
        \caption{Classification error per dataset.}
     \resizebox{\textwidth}{!}{%
    \begin{tabular}{lllllllllllll}
\toprule
\multirow{2}[3]{*}{\bfseries Method} & \multicolumn{2}{c}{\bfseries DBpedia } & \multicolumn{2}{c}{\bfseries  News } & \multicolumn{2}{c}{\bfseries  SetFit } & \multicolumn{2}{c}{\bfseries SST-2} & \multicolumn{2}{c}{\bfseries Tweet} & \multicolumn{2}{c}{\bfseries IMDB} \\
\cmidrule(rl){2-3} \cmidrule(rl){4-5} \cmidrule(rl){6-7} \cmidrule(rl){8-9} \cmidrule(rl){10-11} \cmidrule(rl){12-13}
 & \bfseries 100 \% & \bfseries 10 \%  & \bfseries 100 \%  & \bfseries 10 \%  & \bfseries 100 \%  & \bfseries 10 \%  & \bfseries 100 \%  & \bfseries 10 \%  & \bfseries 100 \%  & \bfseries 10 \%  & \bfseries 100 \%  & \bfseries 10 \%  \\
\midrule
\bfseries Single-Best & 0.0077 & 0.0085 & 0.0462 & 0.0657 & 0.1898 & 0.1338 & 0.0396 & 0.0507 & 0.2012 & 0.2306 & 0.0362 & 0.0455 \\
\bfseries Random-5 & 0.0139 & 0.3157                     & 0.0574        & 0.0833                & 0.2383 & 0.1624                       & 0.0542 & 0.1060       & 0.1925 & 0.2233           & 0.0507 & 0.0657 \\
\bfseries Random-50 & 0.0110 & 0.0082                    & 0.0558        & 0.0786                & 0.1965 & 0.1639                       & 0.0529 & 0.0684       & 0.1898 & 0.2140           & 0.0387 & 0.0497 \\
\bfseries Top-5 & 0.0076 & 0.0077                        & \bfseries 0.0455 & 0.0636             & 0.1846 & 0.1277                       &\bfseries  0.0359 & 0.0488       & 0.1921 & 0.2187           & 0.0328 & \bfseries  0.0416 \\
\bfseries Top-50 & 0.0110 & 0.0083                       & 0.0525        & 0.0651                & 0.1989 & 0.1526                       & 0.0411 & 0.0543         &  0.1885 & 0.2142        & 0.0370 & 0.0446 \\
\bfseries Model Average & 0.0087 & 0.0087               & 0.0533        & 0.0703                & 0.1896 & 0.1450                       & 0.0444 & 0.0564             & 0.1889 & 0.2107         & 0.0392 & 0.0484 \\
\bfseries Greedy-5 & \bfseries 0.0074 & 0.0079           & 0.0459        & 0.0611                & 0.1846 & 0.1261                       & 0.0377 & \bfseries 0.0472 & 0.1953 & 0.2102        & \bfseries0.0321 & 0.0420 \\
\bfseries Greedy-50 & 0.0075 & \bfseries 0.0076           & 0.0459        & \bfseries 0.0593     & \bfseries 0.1843 & \bfseries0.1245    & 0.0376 & 0.0473 & \bfseries 0.1872 &  \bfseries 0.2050 & \bfseries 0.0321 & 0.0420 \\
\bottomrule
\end{tabular}

    }
    \label{tab:error}
\end{table}

\begin{table}[]
    \centering
        \caption{Negative log-likelihood (NLL) per dataset.}
     \resizebox{\textwidth}{!}{%
    \begin{tabular}{lllllllllllll}
\toprule
\multirow{2}[3]{*}{\bfseries Method} & \multicolumn{2}{c}{\bfseries DBpedia } & \multicolumn{2}{c}{\bfseries  News } & \multicolumn{2}{c}{\bfseries  SetFit } & \multicolumn{2}{c}{\bfseries SST-2} & \multicolumn{2}{c}{\bfseries Tweet} & \multicolumn{2}{c}{\bfseries IMDB} \\
\cmidrule(rl){2-3} \cmidrule(rl){4-5} \cmidrule(rl){6-7} \cmidrule(rl){8-9} \cmidrule(rl){10-11} \cmidrule(rl){12-13}
 & \bfseries 100 \% & \bfseries 10 \%  & \bfseries 100 \%  & \bfseries 10 \%  & \bfseries 100 \%  & \bfseries 10 \%  & \bfseries 100 \%  & \bfseries 10 \%  & \bfseries 100 \%  & \bfseries 10 \%  & \bfseries 100 \%  & \bfseries 10 \%  \\
\midrule
\bfseries Single-Best & 0.0497 & 0.0631 & 0.2085 & 0.3369 & 0.8154 & 0.6112 & 0.2037 & 0.2186 & 0.6225 & 0.7870 & 0.1475 & 0.2086 \\
\bfseries Random-5 & 0.0644 & 1.0705 & 0.5032 & 0.4500 & 0.8871 & 0.6488 & 0.2959 & 0.4100 & 0.6763 & 0.6745 & 0.2856 & 0.3051 \\
\bfseries Random-50 & 0.0492 & 0.3900 & 0.5706 & 0.4091 & 0.6728 & 0.6434 & 0.3447 & 0.3788 & 0.6466 & 0.6939 & 0.3483 & 0.3551 \\
\bfseries Top-5 & 0.0424 & 0.0534 & 0.1768 & 0.2423 & 0.7175 & 0.4945 & 0.1468 & 0.2159 & 0.5822 & 0.7060 & 0.1193 & 0.1576 \\
\bfseries Top-50 & 0.0484 & 0.2355 & 0.1796 & 0.2348 & 0.6997 & 0.6379 & 0.1275 & 0.2034 & 0.5181 & 0.7223 & 0.1179 & 0.1320 \\
\bfseries Model Average & 0.0433 & 0.1453 & 0.2461 & 0.2753 & 0.5541 & 0.4602 & 0.1685 & 0.1987 & 0.5143 & 0.5588 & 0.1561 & 0.1716 \\
\bfseries Greedy-5 & 0.0383 & 0.0446 & 0.1751 & 0.2319 & 0.5413 & 0.4037 & 0.1389 & 0.1587 & 0.5085 & 0.5419 & 0.1150 & 0.1272 \\
\bfseries Greedy-50 & \bfseries 0.0358 & \bfseries 0.0364 & \bfseries 0.1582 & \bfseries 0.1978 & \bfseries 0.5290 & \bfseries 0.3572 & \bfseries 0.1167 &\bfseries  0.1365 & \bfseries 0.4769 & \bfseries 0.5077 & \bfseries 0.1031 & \bfseries 0.1241 \\

\bottomrule
\end{tabular}

    }
    \label{tab:nll}
\end{table}

\subsection{Observation 2: Ensembling text classifiers finetuned on 10\% of the training data yields strong results.}

Given the two training splits in the metadataset, we study the advantages of using just $10\%$ of the data for finetuning and post-hoc ensembling. Our results show that, as expected, the best option is to use the whole training data. Nevertheless, we notice that ensembling is also beneficial when training in the subset of data (see Tables \ref{tab:error} and \ref{tab:nll}). Remarkably, ensembling these models sometimes yields better performance than using a single best trained on the whole data. We observe such results for all models under NLL and for two models using \textit{Greedy-50} under the error metrics.

\section{Conclusion}

In this work, we introduced a metadataset containing the predictions of finetuned text classifiers and evaluated common ensembling strategies using this data. Our study provided insights on how simple strategies can improve on top of vanilla single configuration selection in the context of text classification. We empirically showed that even finetuning on small datasets or subsets of data can yield a considerable improvement. Finally, our experiments suggest that the finetuned model and learning rate have an important impact on the final performance.

\section*{Acknowledgments}

The authors gratefully acknowledge the scientific support and HPC resources provided by the Erlangen National High Performance Computing Center (NHR@FAU) of the Friedrich-Alexander-Universität Erlangen-Nürnberg (FAU) under the NHR project v101be. NHR funding is provided by federal and Bavarian state authorities. NHR@FAU hardware is partially funded by the German Research Foundation (DFG) – 440719683. 
This research was partially supported by the following sources: TAILOR, a project funded by EU Horizon 2020 research and innovation programme under GA No 952215; the Deutsche Forschungsgemeinschaft (DFG, German Research Foundation) under grant number 417962828 and 499552394 - SFB 1597; the European Research Council (ERC) Consolidator Grant “Deep Learning 2.0” (grant no. 101045765). Frank Hutter acknowledges financial support by the Hector Foundation. The authors acknowledge support from ELLIS and ELIZA. Funded by the European Union. Views and opinions expressed are however those of the author(s) only and do not necessarily reflect those of the European Union or the ERC. Neither the European Union nor the ERC can be held responsible for them.
\begin{center}\includegraphics[width=0.3\textwidth]{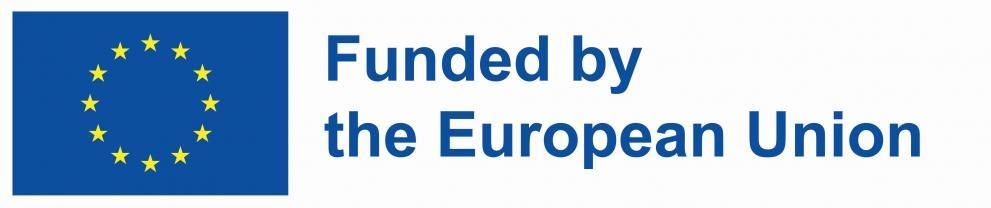}\end{center}. 

\bibliography{references,nes_bib}

\begin{thebibliography}{41}
\providecommand{\natexlab}[1]{#1}
\providecommand{\url}[1]{\texttt{#1}}
\expandafter\ifx\csname urlstyle\endcsname\relax
  \providecommand{\doi}[1]{doi: #1}\else
  \providecommand{\doi}{doi: \begingroup \urlstyle{rm}\Url}\fi

\bibitem[Abburi et~al.(2023)Abburi, Suesserman, Pudota, Veeramani, Bowen, and Bhattacharya]{abduri2027generative}
Abburi, H., Suesserman, M., Pudota, N., Veeramani, B., Bowen, E., and Bhattacharya, S.
\newblock Generative {AI} text classification using ensemble {LLM} approaches.
\newblock In Montes{-}y{-}G{\'{o}}mez, M., Rangel, F., Jim{\'{e}}nez{-}Zafra, S.~M., Casavantes, M., Altuna, B., {\'{A}}lvarez{-}Carmona, M.~{\'{A}}., Bel{-}Enguix, G., Chiruzzo, L., de~la Iglesia, I., Escalante, H.~J., Cumbreras, M. {\'{A}}.~G., Garc{\'{\i}}a{-}D{\'{\i}}az, J.~A., Barba, J. {\'{A}}.~G., Tamayo, R.~L., Lima, S., Moral, P., del Arco, F. M.~P., and Valencia{-}Garc{\'{\i}}a, R. (eds.), \emph{Proceedings of the Iberian Languages Evaluation Forum (IberLEF 2023) co-located with the Conference of the Spanish Society for Natural Language Processing {(SEPLN} 2023), Ja{\'{e}}n, Spain, September 26, 2023}, volume 3496 of \emph{{CEUR} Workshop Proceedings}. CEUR-WS.org, 2023.

\bibitem[Arango et~al.(2023)Arango, Ferreira, Kadra, Hutter, and Grabocka]{arango2023quick}
Arango, S.~P., Ferreira, F., Kadra, A., Hutter, F., and Grabocka, J.
\newblock Quick-tune: Quickly learning which pretrained model to finetune and how.
\newblock \emph{arXiv preprint arXiv:2306.03828}, 2023.

\bibitem[Blundell et~al.(2015)Blundell, Cornebise, Kavukcuoglu, and Wierstra]{pmlr-v37-blundell15}
Blundell, C., Cornebise, J., Kavukcuoglu, K., and Wierstra, D.
\newblock Weight uncertainty in neural network.
\newblock In \emph{Proceedings of the 32nd International Conference on Machine Learning}, volume~37 of \emph{Proceedings of Machine Learning Research}, pp.\  1613--1622, Lille, France, 07--09 Jul 2015. PMLR.
\newblock URL \url{http://proceedings.mlr.press/v37/blundell15.html}.

\bibitem[Borchert et~al.(2022)Borchert, Salinas, Flunkert, Januschowski, and G{\"u}nnemann]{borchert2022multi}
Borchert, O., Salinas, D., Flunkert, V., Januschowski, T., and G{\"u}nnemann, S.
\newblock Multi-objective model selection for time series forecasting.
\newblock \emph{arXiv preprint arXiv:2202.08485}, 2022.

\bibitem[Caruana et~al.(2004)Caruana, Niculescu-Mizil, Crew, and Ksikes]{caruana2004ensemble}
Caruana, R., Niculescu-Mizil, A., Crew, G., and Ksikes, A.
\newblock Ensemble selection from libraries of models.
\newblock In \emph{Proceedings of the twenty-first international conference on Machine learning}, pp.\ ~18, 2004.

\bibitem[Caruana et~al.(2006)Caruana, Munson, and Niculescu-Mizil]{caruana2006getting}
Caruana, R., Munson, A., and Niculescu-Mizil, A.
\newblock Getting the most out of ensemble selection.
\newblock In \emph{Sixth International Conference on Data Mining (ICDM'06)}, pp.\  828--833. IEEE, 2006.

\bibitem[Dettmers et~al.(2024)Dettmers, Pagnoni, Holtzman, and Zettlemoyer]{dettmers2024qlora}
Dettmers, T., Pagnoni, A., Holtzman, A., and Zettlemoyer, L.
\newblock Qlora: Efficient finetuning of quantized llms.
\newblock \emph{Advances in Neural Information Processing Systems}, 36, 2024.

\bibitem[Devlin et~al.(2018)Devlin, Chang, Lee, and Toutanova]{DBLP:journals/corr/abs-1810-04805}
Devlin, J., Chang, M., Lee, K., and Toutanova, K.
\newblock {BERT:} pre-training of deep bidirectional transformers for language understanding.
\newblock \emph{CoRR}, abs/1810.04805, 2018.
\newblock URL \url{http://arxiv.org/abs/1810.04805}.

\bibitem[Dietterich(2000)]{dietterich_ensemble_methods}
Dietterich, T.~G.
\newblock Ensemble {Methods} in {Machine} {Learning}.
\newblock In \emph{Multiple Classifier Systems}, pp.\  1--15, Berlin, Heidelberg, 2000. Springer Berlin Heidelberg.
\newblock ISBN 978-3-540-45014-6.

\bibitem[Erickson et~al.(2020)Erickson, Mueller, Shirkov, Zhang, Larroy, Li, and Smola]{agtabular}
Erickson, N., Mueller, J., Shirkov, A., Zhang, H., Larroy, P., Li, M., and Smola, A.
\newblock Autogluon-tabular: Robust and accurate automl for structured data.
\newblock \emph{arXiv preprint arXiv:2003.06505}, 2020.

\bibitem[Gal \& Ghahramani(2016)Gal and Ghahramani]{pmlr-v48-gal16}
Gal, Y. and Ghahramani, Z.
\newblock Dropout as a bayesian approximation: Representing model uncertainty in deep learning.
\newblock In \emph{Proceedings of the 33rd International Conference on Machine Learning}, volume~48 of \emph{Proceedings of Machine Learning Research}, pp.\  1050--1059, New York, New York, USA, 20--22 Jun 2016. PMLR.
\newblock URL \url{http://proceedings.mlr.press/v48/gal16.html}.

\bibitem[Gustafsson et~al.(2020)Gustafsson, Danelljan, and Sch{\"o}n]{gustafsson2019evaluating}
Gustafsson, F.~K., Danelljan, M., and Sch{\"o}n, T.~B.
\newblock Evaluating {Scalable} {Bayesian} {Deep} {Learning} {Methods} for {Robust} {Computer} {Vision}.
\newblock In \emph{The IEEE Conference on Computer Vision and Pattern Recognition (CVPR) Workshops}, June 2020.

\bibitem[Han et~al.(2024)Han, Gao, Liu, Zhang, and Zhang]{han2024peftsurvey}
Han, Z., Gao, C., Liu, J., Zhang, J., and Zhang, S.~Q.
\newblock Parameter-efficient fine-tuning for large models: A comprehensive survey, 2024.
\newblock URL \url{https://arxiv.org/abs/2403.14608}.

\bibitem[{Hansen} \& {Salamon}(1990){Hansen} and {Salamon}]{hansen_ensembles}
{Hansen}, L.~K. and {Salamon}, P.
\newblock Neural {network} {ensembles}.
\newblock \emph{IEEE Transactions on Pattern Analysis and Machine Intelligence}, 12\penalty0 (10):\penalty0 993--1001, 1990.

\bibitem[Houlsby et~al.(2019)Houlsby, Giurgiu, Jastrzebski, Morrone, De~Laroussilhe, Gesmundo, Attariyan, and Gelly]{houlsby19adapters}
Houlsby, N., Giurgiu, A., Jastrzebski, S., Morrone, B., De~Laroussilhe, Q., Gesmundo, A., Attariyan, M., and Gelly, S.
\newblock Parameter-efficient transfer learning for {NLP}.
\newblock In Chaudhuri, K. and Salakhutdinov, R. (eds.), \emph{Proceedings of the 36th International Conference on Machine Learning}, volume~97 of \emph{Proceedings of Machine Learning Research}, pp.\  2790--2799. PMLR, 09--15 Jun 2019.
\newblock URL \url{https://proceedings.mlr.press/v97/houlsby19a.html}.

\bibitem[Howard \& Ruder(2018)Howard and Ruder]{ruder2018universal}
Howard, J. and Ruder, S.
\newblock Universal language model fine-tuning for text classification.
\newblock In Gurevych, I. and Miyao, Y. (eds.), \emph{Proceedings of the 56th Annual Meeting of the Association for Computational Linguistics, {ACL} 2018, Melbourne, Australia, July 15-20, 2018, Volume 1: Long Papers}, pp.\  328--339. Association for Computational Linguistics, 2018.

\bibitem[Hu et~al.(2022)Hu, Shen, Wallis, Allen-Zhu, Li, Wang, Wang, and Chen]{hu2022lora}
Hu, E.~J., Shen, Y., Wallis, P., Allen-Zhu, Z., Li, Y., Wang, S., Wang, L., and Chen, W.
\newblock Lo{RA}: Low-rank adaptation of large language models.
\newblock In \emph{International Conference on Learning Representations}, 2022.
\newblock URL \url{https://openreview.net/forum?id=nZeVKeeFYf9}.

\bibitem[Koh et~al.(2021)Koh, Sagawa, Marklund, Xie, Zhang, Balsubramani, Hu, Yasunaga, Phillips, Gao, Lee, David, Stavness, Guo, Earnshaw, Haque, Beery, Leskovec, Kundaje, Pierson, Levine, Finn, and Liang]{wilds-koh21a}
Koh, P.~W., Sagawa, S., Marklund, H., Xie, S.~M., Zhang, M., Balsubramani, A., Hu, W., Yasunaga, M., Phillips, R.~L., Gao, I., Lee, T., David, E., Stavness, I., Guo, W., Earnshaw, B., Haque, I., Beery, S.~M., Leskovec, J., Kundaje, A., Pierson, E., Levine, S., Finn, C., and Liang, P.
\newblock Wilds: A benchmark of in-the-wild distribution shifts.
\newblock In \emph{Proceedings of the 38th International Conference on Machine Learning}, volume 139 of \emph{Proceedings of Machine Learning Research}, pp.\  5637--5664. PMLR, 18--24 Jul 2021.

\bibitem[Krogh \& Vedelsby(1995)Krogh and Vedelsby]{krogh-ens}
Krogh, A. and Vedelsby, J.
\newblock Neural {Network} {Ensembles}, {Cross} {Validation}, and {Active} {Learning}.
\newblock In \emph{Advances in Neural Information Processing Systems 7}, pp.\  231--238. MIT Press, 1995.

\bibitem[Lakshminarayanan et~al.(2017)Lakshminarayanan, Pritzel, and Blundell]{lakshminarayanan2017simple}
Lakshminarayanan, B., Pritzel, A., and Blundell, C.
\newblock Simple and scalable predictive uncertainty estimation using deep ensembles.
\newblock \emph{Advances in neural information processing systems}, 30, 2017.

\bibitem[Lan et~al.(2019)Lan, Chen, Goodman, Gimpel, Sharma, and Soricut]{albert}
Lan, Z., Chen, M., Goodman, S., Gimpel, K., Sharma, P., and Soricut, R.
\newblock {ALBERT:} {A} lite {BERT} for self-supervised learning of language representations.
\newblock \emph{CoRR}, abs/1909.11942, 2019.
\newblock URL \url{http://arxiv.org/abs/1909.11942}.

\bibitem[Lewis et~al.(2019)Lewis, Liu, Goyal, Ghazvininejad, Mohamed, Levy, Stoyanov, and Zettlemoyer]{bart}
Lewis, M., Liu, Y., Goyal, N., Ghazvininejad, M., Mohamed, A., Levy, O., Stoyanov, V., and Zettlemoyer, L.
\newblock {BART:} denoising sequence-to-sequence pre-training for natural language generation, translation, and comprehension.
\newblock \emph{CoRR}, abs/1910.13461, 2019.
\newblock URL \url{http://arxiv.org/abs/1910.13461}.

\bibitem[Maas et~al.(2011)Maas, Daly, Pham, Huang, Ng, and Potts]{imdb_dataset}
Maas, A.~L., Daly, R.~E., Pham, P.~T., Huang, D., Ng, A.~Y., and Potts, C.
\newblock Learning word vectors for sentiment analysis.
\newblock In \emph{Proceedings of the 49th Annual Meeting of the Association for Computational Linguistics: Human Language Technologies}, pp.\  142--150, Portland, Oregon, USA, June 2011. Association for Computational Linguistics.
\newblock URL \url{http://www.aclweb.org/anthology/P11-1015}.

\bibitem[Maggie(2020)]{tweet-sentiment-extraction}
Maggie, Phil~Culliton, W.~C.
\newblock Tweet sentiment extraction, 2020.
\newblock URL \url{https://kaggle.com/competitions/tweet-sentiment-extraction}.

\bibitem[Mendes et~al.(2012)Mendes, Jakob, and Bizer]{mendes2012dbpedia}
Mendes, P.~N., Jakob, M., and Bizer, C.
\newblock \emph{DBpedia: A multilingual cross-domain knowledge base}.
\newblock European Language Resources Association (ELRA), 2012.

\bibitem[Nangia et~al.(2017)Nangia, Williams, Lazaridou, and Bowman]{nangia2017repeval}
Nangia, N., Williams, A., Lazaridou, A., and Bowman, S.~R.
\newblock The repeval 2017 shared task: Multi-genre natural language inference with sentence representations.
\newblock \emph{arXiv preprint arXiv:1707.08172}, 2017.

\bibitem[Ovadia et~al.(2019)Ovadia, Fertig, Ren, Nado, Sculley, Nowozin, Dillon, Lakshminarayanan, and Snoek]{ovadia19}
Ovadia, Y., Fertig, E., Ren, J., Nado, Z., Sculley, D., Nowozin, S., Dillon, J., Lakshminarayanan, B., and Snoek, J.
\newblock Can you trust your model\textquotesingle s uncertainty? {Evaluating} predictive uncertainty under dataset shift.
\newblock In \emph{Advances in Neural Information Processing Systems 32}, pp.\  13991--14002. Curran Associates, Inc., 2019.

\bibitem[Purucker \& Beel(2022)Purucker and Beel]{purucker2022assembled}
Purucker, L.~O. and Beel, J.
\newblock Assembled-openml: Creating efficient benchmarks for ensembles in automl with openml.
\newblock In \emph{First Conference on Automated Machine Learning (Late-Breaking Workshop)}, 2022.

\bibitem[Purucker \& Beel(2023)Purucker and Beel]{purucker2023cma}
Purucker, L.~O. and Beel, J.
\newblock Cma-es for post hoc ensembling in automl: A great success and salvageable failure.
\newblock In \emph{International Conference on Automated Machine Learning}, pp.\  1--1. PMLR, 2023.

\bibitem[Purucker et~al.(2023)Purucker, Schneider, Anastacio, Beel, Bischl, and Hoos]{purucker2023q}
Purucker, L.~O., Schneider, L., Anastacio, M., Beel, J., Bischl, B., and Hoos, H.
\newblock Q(d)o-es: Population-based quality (diversity) optimisation for post hoc ensemble selection in automl.
\newblock In \emph{International Conference on Automated Machine Learning}, pp.\  10--1. PMLR, 2023.

\bibitem[Radford et~al.(2018)Radford, Narasimhan, Salimans, and Sutskever]{radford2018plm}
Radford, A., Narasimhan, K., Salimans, T., and Sutskever, I.
\newblock Improving language understanding with unsupervised learning.
\newblock Technical Report, OpenAI, 2018.
\newblock URL \url{https://openai.com/index/language-unsupervised/}.

\bibitem[Radford et~al.(2019)Radford, Wu, Child, Luan, Amodei, and Sutskever]{radford2019language}
Radford, A., Wu, J., Child, R., Luan, D., Amodei, D., and Sutskever, I.
\newblock Language models are unsupervised multitask learners.
\newblock 2019.

\bibitem[Raffel et~al.(2020)Raffel, Shazeer, Roberts, Lee, Narang, Matena, Zhou, Li, and Liu]{2020t5}
Raffel, C., Shazeer, N., Roberts, A., Lee, K., Narang, S., Matena, M., Zhou, Y., Li, W., and Liu, P.~J.
\newblock Exploring the limits of transfer learning with a unified text-to-text transformer.
\newblock \emph{Journal of Machine Learning Research}, 21\penalty0 (140):\penalty0 1--67, 2020.
\newblock URL \url{http://jmlr.org/papers/v21/20-074.html}.

\bibitem[Salinas \& Erickson(2023)Salinas and Erickson]{salinas2023tabrepo}
Salinas, D. and Erickson, N.
\newblock Tabrepo: A large scale repository of tabular model evaluations and its automl applications.
\newblock \emph{arXiv preprint arXiv:2311.02971}, 2023.

\bibitem[Seligmann et~al.(2024)Seligmann, Becker, Volpp, and Neumann]{beyondDE_23}
Seligmann, F., Becker, P., Volpp, M., and Neumann, G.
\newblock Beyond deep ensembles: a large-scale evaluation of bayesian deep learning under distribution shift.
\newblock In \emph{Proceedings of the 37th International Conference on Neural Information Processing Systems}, NeurIPS '23, Red Hook, NY, USA, 2024. Curran Associates Inc.

\bibitem[Socher et~al.(2013)Socher, Perelygin, Wu, Chuang, Manning, Ng, and Potts]{socher-etal-2013-recursive}
Socher, R., Perelygin, A., Wu, J., Chuang, J., Manning, C.~D., Ng, A., and Potts, C.
\newblock Recursive deep models for semantic compositionality over a sentiment treebank.
\newblock In \emph{Proceedings of the 2013 Conference on Empirical Methods in Natural Language Processing}, pp.\  1631--1642, Seattle, Washington, USA, October 2013. Association for Computational Linguistics.
\newblock URL \url{https://www.aclweb.org/anthology/D13-1170}.

\bibitem[Sung et~al.(2021)Sung, Nair, and Raffel]{sung2021subnets}
Sung, Y.-L., Nair, V., and Raffel, C.~A.
\newblock Training neural networks with fixed sparse masks.
\newblock \emph{Advances in Neural Information Processing Systems}, 34:\penalty0 24193--24205, 2021.

\bibitem[Tunstall et~al.(2021)Tunstall, Pereg, Bates, Wasserblat, Eun, Korat, Reimers, and Aarsen]{setfit-mnli}
Tunstall, L., Pereg, O., Bates, L., Wasserblat, M., Eun, U., Korat, D., Reimers, N., and Aarsen, T.
\newblock Setfit-mnli, 2021.
\newblock URL \url{https://huggingface.co/datasets/SetFit/mnli}.

\bibitem[Tunstall et~al.(2022)Tunstall, Reimers, Jo, Bates, Korat, Wasserblat, and Pereg]{setfit_paper}
Tunstall, L., Reimers, N., Jo, U. E.~S., Bates, L., Korat, D., Wasserblat, M., and Pereg, O.
\newblock Efficient few-shot learning without prompts, 2022.
\newblock URL \url{https://arxiv.org/abs/2209.11055}.

\bibitem[Welling \& Teh(2011)Welling and Teh]{sgld11}
Welling, M. and Teh, Y.~W.
\newblock Bayesian learning via stochastic gradient langevin dynamics.
\newblock In \emph{Proceedings of the 28th International Conference on International Conference on Machine Learning}, ICML'11, pp.\  681–688, Madison, WI, USA, 2011. Omnipress.
\newblock ISBN 9781450306195.

\bibitem[Zhang et~al.(2015)Zhang, Zhao, and LeCun]{Zhang2015CharacterlevelCN}
Zhang, X., Zhao, J., and LeCun, Y.
\newblock Character-level convolutional networks for text classification.
\newblock In Cortes, C., Lawrence, N., Lee, D., Sugiyama, M., and Garnett, R. (eds.), \emph{Advances in Neural Information Processing Systems}, volume~28. Curran Associates, Inc., 2015.
\newblock URL \url{https://proceedings.neurips.cc/paper_files/paper/2015/file/250cf8b51c773f3f8dc8b4be867a9a02-Paper.pdf}.

\end{thebibliography}
\bibliographystyle{neurips_fitml_2024}

\clearpage
\appendix

\section{Details on Datasets}
\label{apx:dataset_information}

\paragraph{IMDB~\citep{imdb_dataset}}
The IMDB dataset contains reviews for movies and their binary sentiment. 
We only use the labeled training and test data. 
The data source we used is \url{https://huggingface.co/datasets/stanfordnlp/imdb}.

\paragraph{Tweet~\citep{tweet-sentiment-extraction}}
The Tweet dataset contains the text of tweets and their sentiment label.
The data was initially curated for a Kaggle competition.
The data source we used is \url{https://www.kaggle.com/competitions/tweet-sentiment-extraction}.

\paragraph{DBpedia and News~\citep{Zhang2015CharacterlevelCN}}
The DBpedia and News datasets were created by \citet{Zhang2015CharacterlevelCN} for benchmarking deep learning models for text classification tasks.

We use the AG's News dataset, consisting of the title and description fields of news articles from the web. The data source we used is \url{https://huggingface.co/datasets/fancyzhx/ag_news}.

The DBpedia dataset contains the title and abstract of Wikipedia articles sourced from DBpedia 2014 \citep{mendes2012dbpedia}.
The data source we used is \url{https://huggingface.co/datasets/fancyzhx/dbpedia_14}.

\paragraph{SST-2~\citep{socher-etal-2013-recursive}}
The Stanford Sentiment Treebank with two classes (SST-2) is a corpus of individual sentences from movie reviews. 
Three human judges labeled the sentences as having (somewhat) negative or (somewhat) positive sentiments.
The data source we used is \url{https://huggingface.co/datasets/stanfordnlp/sst2}.

\paragraph{SetFit~\citep{setfit-mnli}}
Lastly, we use the SetFit \citep{setfit_paper} version of the Multi-Genre Natural Language Inference (MNLI) corpus \citep{nangia2017repeval} as a dataset. 
The corpus encompasses text pairs from various sources, such as transcribed speech or fiction.
Each text pair is labeled with whether one text entails the other, contradicts the other, or if they are neutral to each other. 
The data source we used is \url{https://huggingface.co/datasets/SetFit/mnli}.

\section{Additional Results}

\begin{figure}[H]
    \centering
    \includegraphics[width=0.5\linewidth]{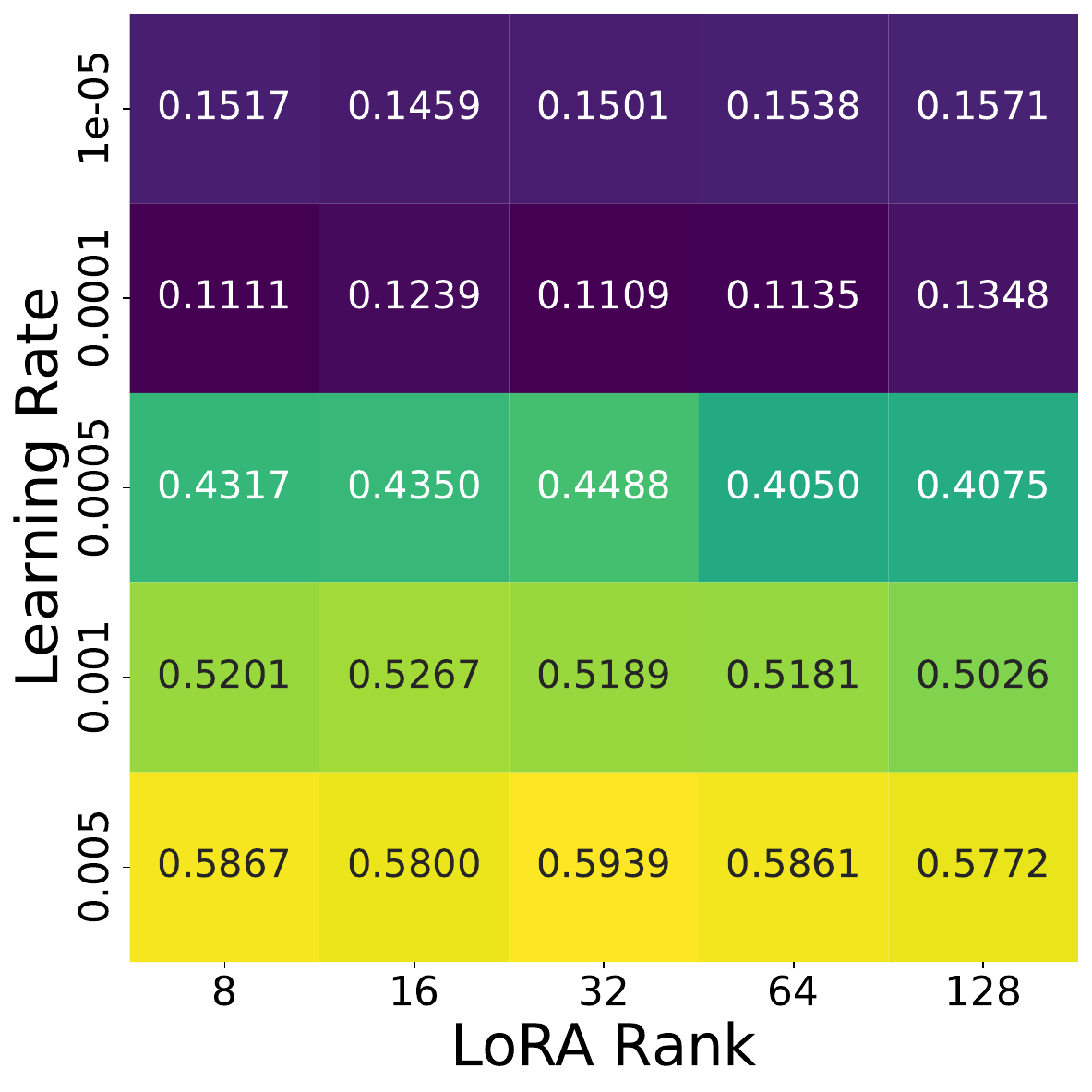}
    \caption{Error for different hyperparameters using 10 \% of the data.}
    \label{fig:error_hyper_mini}
\end{figure}

\begin{figure}[H]
    \centering
    \includegraphics[width=1.\linewidth]{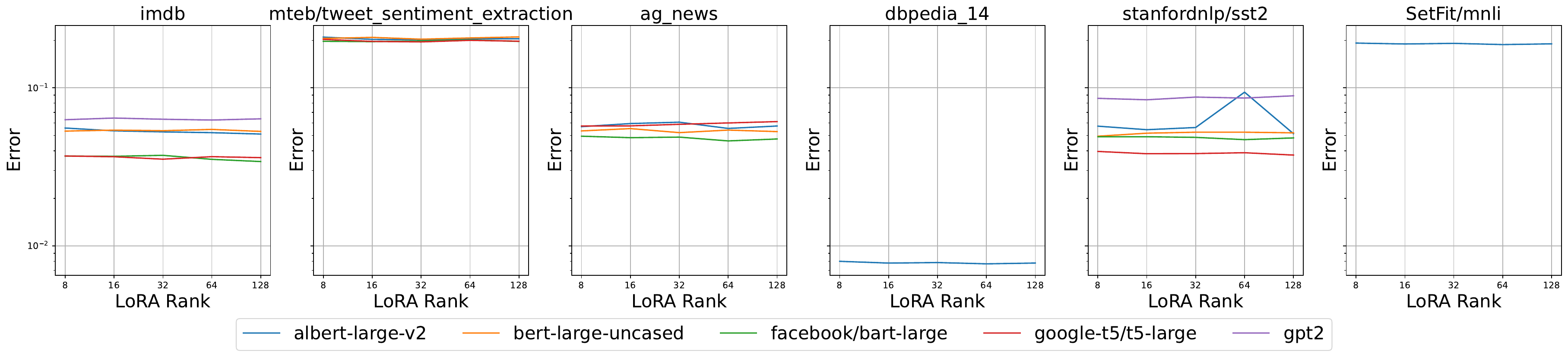}
    \caption{Error vs. LoRA Rank, \textit{extended} version. The error variation is small across different LoRA rank values.}
    \label{fig:error_lora_extended}
\end{figure}

We present additional results:
\begin{itemize}
    \item Mean error for different hyperparameters using a subset of data in Figure \ref{fig:error_hyper_mini}.
    \item Finetuning time for every dataset in Table \ref{tab:times}.
    \item Comparison performance per model in Table \ref{tab:error_per_mode}.
    \item Comparison of different values of LoRA rank dimension in Figures \ref{fig:error_lora_extended}
\end{itemize}

\begin{table}[H]
    \centering
    \caption{Training times per dataset.}
    \begin{tabular}{lcccc}
\toprule
 & \multicolumn{2}{c}{\bfseries Average (Min.)} & \multicolumn{2}{c}{\bfseries Total (Hrs.)} \\ \cmidrule(lr){2-3} \cmidrule(lr){4-5}
& \bfseries Extended & \bfseries Mini & \bfseries Extended & \bfseries Mini \\
\midrule
\bfseries Set-Fit & 104.49 & 24.32 & 217.6963& 405.4354 \\
\bfseries News & 91.6443 & 12.20 & 756.0661 & 244.1131 \\
\bfseries DBPedia & 186.22 & 36.99 & 387.9752 & 400.8265 \\
\bfseries IMDB & 26.84 & 2.71 & 279.64 & 56.47 \\
\bfseries Tweet & 34.54 & 3.46 & 287.83& 57.77 \\
\bfseries SST2 & 57.97 & 5.79 & 603.94 & 120.63 \\
\bottomrule
\end{tabular}
    \label{tab:times}
\end{table}

\begin{table}[H]
    \caption{Error per Model.}
    \centering
     \resizebox{\textwidth}{!}{%
    \begin{tabular}{lcccccccccccc}
\toprule
\multirow{2}[3]{*}{\bfseries Method} & \multicolumn{2}{c}{\bfseries IMDB } & \multicolumn{2}{c}{\bfseries  Tweet } & \multicolumn{2}{c}{\bfseries  News } & \multicolumn{2}{c}{\bfseries DBpedia} & \multicolumn{2}{c}{\bfseries SST2} & \multicolumn{2}{c}{\bfseries Set-Fit} \\
\cmidrule(rl){2-7} \cmidrule(lr){8-13}
 & \bfseries 100 \% & \bfseries 10 \%  & \bfseries 100 \%  & \bfseries 10 \%  & \bfseries 100 \%  & \bfseries 10 \%  & \bfseries 100 \%  & \bfseries 10 \%  & \bfseries 100 \%  & \bfseries 10 \%  & \bfseries 100 \%  & \bfseries 10 \%  \\
\midrule

\bfseries GPT2 & 0.0576 & 0.0817                    & -             & -      &  0.0611   & 0.0736   & \bfseries0.0077   & 0.0103    & 0.0840  & 0.1174 & \bfseries0.1898 & 0.2388  \\
\bfseries Bert-Large & 0.0540 & 0.0752               & 0.2031    & 0.2365   & 0.0540    &  0.0772   & -         & \bfseries 0.0085    & 0.0516 & 0.0809  & - & 0.2007 \\
\bfseries Albert-Large & 0.0534 & 0.0650            & 0.2043     & 0.2439   & 0.0553     & 0.0807   & -         & 0.0105    & 0.0513 & 0.0917 & - & 0.1901 \\
\bfseries Bart-Large & \bfseries 0.0342 & 0.0459    & 0.2011     & 0.2306   & \bfseries 0.0461    & \bfseries0.0656  & -        & -         & 0.0482 & 0.0654 & - & \bfseries 0.1337 \\
\bfseries T5-Large & 0.0362  & \bfseries 0.0455     & \bfseries 0.1972  & \bfseries 0.2303 & -           & 0.0735     & -       & -         & \bfseries 0.0396&  \bfseries0.0506 & - & - \\
\bottomrule
\end{tabular}

    }
    \label{tab:error_per_mode}
\end{table}

\begin{figure}[H]
    \centering
    \includegraphics[width=1.\linewidth]{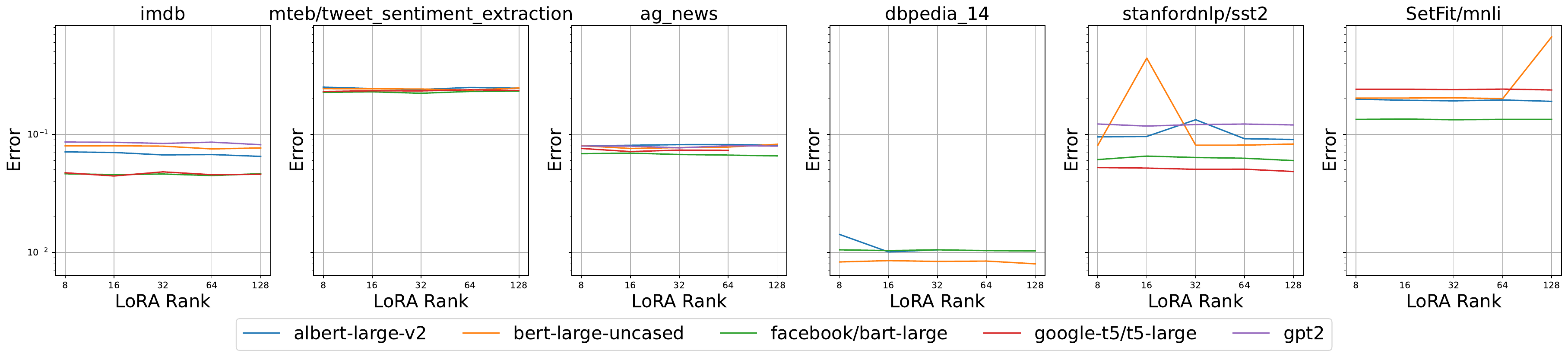}
    \caption{Error vs. LoRA Rank, \textit{mini} version. The error variation is small across different LoRA rank values.}
    \label{fig:error_lora_mini}
\end{figure}

\end{document}